\documentclass[runningheads]{llncs}
\usepackage{graphicx}
\usepackage{amsmath}
\usepackage{hyperref}
\usepackage{url}
\usepackage{xcolor}
\usepackage[utf8]{inputenc}
\usepackage{multirow}
\usepackage[T2A]{fontenc}
\usepackage[russian,english]{babel}
\usepackage{subcaption}
\usepackage{comment}
\usepackage{enumerate}
\usepackage{paralist}

\author{Ekaterina Artemova\inst{3} \and 
Tatiana Batura\inst{4,7} \and 
Anna Golenkovskaya\inst{6} \and 
Vitaly Ivanin\inst{1,2} \and 
Vladimir Ivanov\inst{5} \and 
Veronika Sarkisyan\inst{3} \and 
Ivan Smurov \inst{1,2} \and
Elena Tutubalina\inst{3} 
}
\authorrunning{V. Ivanin et al.}
\institute{ABBYY, Russia \\
\email{\{vitalii.ivanin,ivan.smurov\}@abbyy.com}
\and
Moscow Institute of Physics and Technology, Russia 
\and
National Research University Higher School of Economics, Russia \\
\and
Novosibirsk State University, Russia
\and
Innopolis University, Russia
\and
Kazan Federal University, Russia
\and
A.P. Ershov Institute of Informatics Systems SB RAS, Russia}

\authorrunning{}

\begin{document}
\title{So What's the Plan? Mining Strategic Planning Documents}

\maketitle              % typeset the header of the contribution
\begin{abstract}
In this paper we present a corpus of Russian strategic planning documents, RuREBus. This project is grounded both from language technology and e-government perspectives. Not only new language sources and tools are being developed, but also their applications to e-goverment research.

We demonstrate the pipeline for creating a text corpus  from scratch. First, the annotation schema is designed. Next texts are marked up using human-in-the-loop strategy, so that preliminary annotations are derived from a machine learning model and are manually corrected. 

The amount of annotated texts is large enough to showcase what insights can be gained from RuREBus. 

\end{abstract}

\section{Introduction}
 Each Russian federal and municipal subject publishes several strategic planning documents per year, related to various directions of region development such as medicine, education, ecology, etc. The tasks, presented in the strategic planning documents, should meet several criteria, such as to fit into the budget and be economically sensible, be aligned with the state and local policy and satisfy the population's expectations. Modern language technology helps to assess strategic planning documents and to gain insights into the general directions of development. 
 
\begin{figure}
    \centering
    \includegraphics[width = 0.6\textwidth]{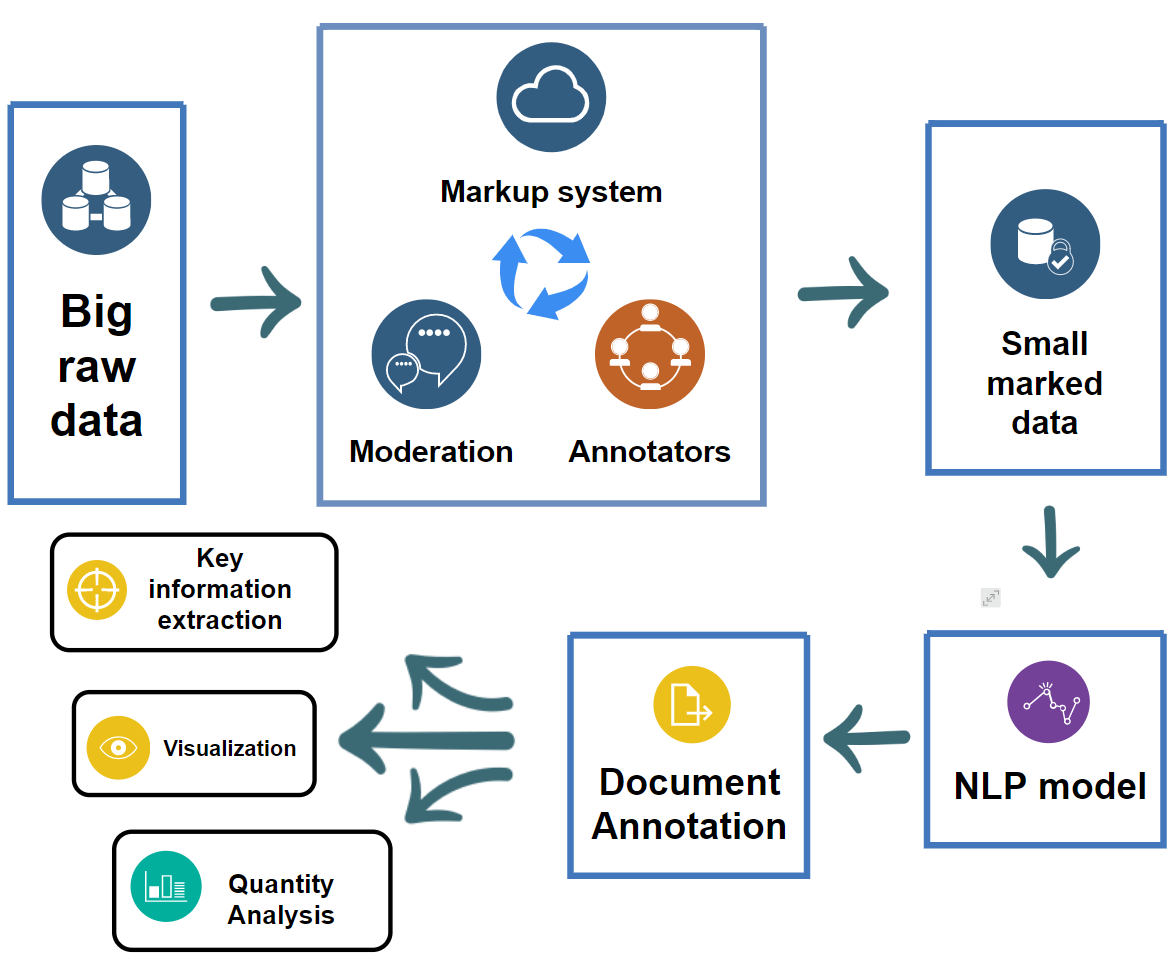}
    \caption{Common NLP pipeline.}
    \label{fig:ppln}
\end{figure}

 In this project, we intend to demonstrate the potential and the value of text-driven analysis when applied to strategic planning. Our approach exploits a common NLP pipeline, which involves annotation of large amount texts and training machine learning models. The most important steps of the pipeline are shown in Figure \ref{fig:ppln}. This pipeline is justified by recent success in other applications, such as processing of clinical stories \cite{holderness2019proceedings} or statements of claim \cite{dale2019law}. Not only we annotate a large amount of texts according to an in-house annotation scheme, but also we use a number of supervised techniques, which allow to extract high-quality information from the documents and convey a more specific and comprehensive picture of the strategic planning and its significance.

 The government documents have not been widely subjected to processing and analysis. This means that we face a need to develop the whole domain-specific pipeline of annotation, information extraction and pre-training of language models. We showcase language technology capabilities. We present an annotation schema to markup named entities and relations,  exploit active learning to annotate hundreds of documents and use state of the art methods for named entity recognition and relation extraction \textbf{to facilitate manual annotation}.
 
 The contributions of the project are threefold. First, a new dataset is being developed, which can be used both by computer science and economics communities for further studies. Second, a number of tools for processing documents in Russian will be released. Third, the dataset will help to conduct detailed analysis of strategic documents, to compare federal subjects and administrative districts in terms of their goals and budget requirement. 
 
%  \textcolor{red}{VI: I would omit or rewrite this one:} Third, we hope to achieve meaningful results for the domain.
 
% Currently, our project is in its early stages and this paper is more of a work-in-progress side. 
% \bibliography{references} 
% \bibliographystyle{splncs04}

\section{Why we build the corpus} 

The annotation schema we have developed for this project provides a powerful tool for strategic document analysis. Indeed listing its possible applications for the domain is one of the main contributions of this paper (see Sections 5-6 for details). However, what is arguably even more important from pure natural language processing perspective is that our dataset can be used as a more fitting case study for structuring unstructured information than other existing datasets. This is a rather bold claim that we intend to argue for in this section.

Structuring unstructured information or to be more specific converting data from text form into database-friendly (i. e. table) form is one of the most popular NLP business applications. Standard techniques used in order to solve this task are named entity recognition(or NER) and relation extraction (or RE). Both NER and RE are well-studied and there exist popular academic benchmarks for both tasks (CoNLL-2003 \cite{tjong-kim-sang-de-meulder-2003-introduction}, ACE-2004 and ACE-2005 \cite{doddington-etal-2004-automatic,ace04,ace05}, SemEval-2010 Task 8 \cite{hendrickx-etal-2010-semeval}, FactRuEval-2016 \cite{f4aa6d516fa74e5992076aa5d3b2a5f1}). There are, however, several important differences between any of aforementioned benchmarks and a typical business case dataset, the most important one being as follows.

Business case texts are usually domain-specific (e. g. legal) texts that can contain less than perfect language or other irregularities (ponderous sentences with complicated syntactic structure, slang etc.). Academic baselines, on the other hand, typically consist of well-written news or biography texts without any irregularities of this kind.

To sum up a popular perception that NER (and to lesser extent RE) is basically a solved task can potentially be to a large extent a product of existing academic benchmarks. While recent years have provided for several major breakthroughs in these tasks, results one can obtain on real-world client corpora are often much more modest than ones reported by scholars \cite{lin-etal-2019-reliability}. 

Given these considerations we decided to create a corpus closer to industrial NER and RE implementation than existing academic ones. Our corpus consists of unadopted domain-specific texts with many irregularities that can be found in practical applications. Hopefully, it can be a better benchmark for checking the suitability of a particular NER and RE model to business scenarios. 

While this particular use of our corpus is not in the main focus of this paper, it was difficult for us to not provide one of our key motivations to create it.

\section{Related work} 

To the best of our knowledge, there are little NLP applications to the e-Goverment domain in general and strategic planning in particular. \cite{alekseychuk2019processing} present the only project of the unsupervised analysis of strategic planning documents. Other e-Goverment applications include processing country statements, governmental web-sites, e-petitions and other social media sources. 

\subsection{Processing e-government documents}

NLP methods allow to extract and structure information of governmental activity. Baturo and Dasandi \cite{UN} used topic modeling to analyze the agenda-setting process of the United Nations based on the UN General Debate corpus \cite{dataset} consisting of over 7300 country statements from 1970 to 2014. In \cite{China} Shen et al. explored Web data and government websites in Beijing, Shanghai, Wuhan, Guangzhou and Chengdu to conduct comparative analysis on the development of the five metropolia e-governments. Albarghothi et al. \cite{arabic} introduced an Automatic Extraction Dataset System (AEDS) tool that constructs an ontology-based Semantic Web from Arabic web pages related to Dubai's e-government services. The system automatically extracts textual data from the website, detects keywords, and finally maps the page to ontology via Protégé tool.

\subsection{Processing petitions}

NLP methods are widely used to aggregate and summarize public opinion, expressed in the form of electronic petitions.
The concept of e-democracy implicates open communication between government and citizens, which in most cases involves the processing of a large amount of unstructured textual information \cite{India}. Rao and Dey describe the scheme of citizens’ and stakeholders’ participation in Indian e-governance which allows the government to collect feedback from citizens and correct policies and acts according to it.

Evangelopoulos and Visinescu \cite{US_people} analyze appeals to the U.S. government, in particular, SMS messages from Africans, sent during Barack Obama's visit to Ghana in July 2009 and data from SAVE Award  - initiative, aiming to make the U.S. government more effective and efficient at spending taxpayers' money. For each of the corpus, authors extracted key topics with Latent Semantic Analysis (LSA) to explore trends in public opinion.

Suh et al. \cite{Korea} applied keyword extraction algorithms based on $tf-idf$ and $k$-means clustering to detect and track petitions groups on Korean e-People petition portal. To forecast the future petition trends, radial basis function neural networks were us.

\subsection{Processing Russian documents}

Metsker et al. \cite{metsker2019natural} process almost 30 million of Russian court decisions to estimate the effectiveness of  legislative change and to identify regional features of law enforcement. Alekseychuk et al. \cite{alekseychuk2019processing} use such unsupervised techniques as topic modelling and word embeddings to induce a taxonomy of regional strategic goals, extracted from strategic documents. This study motivated us to deep into the corpus of strategic documents. Our work extends this project, as we annotate the corpus with a detailed relation scheme. This allows of more detailed analysis.

\section{Corpus annotation pipeline} 

\subsection{Annotation Guidelines} 
\begin{figure}
    \centering
    \includegraphics[width=\textwidth]{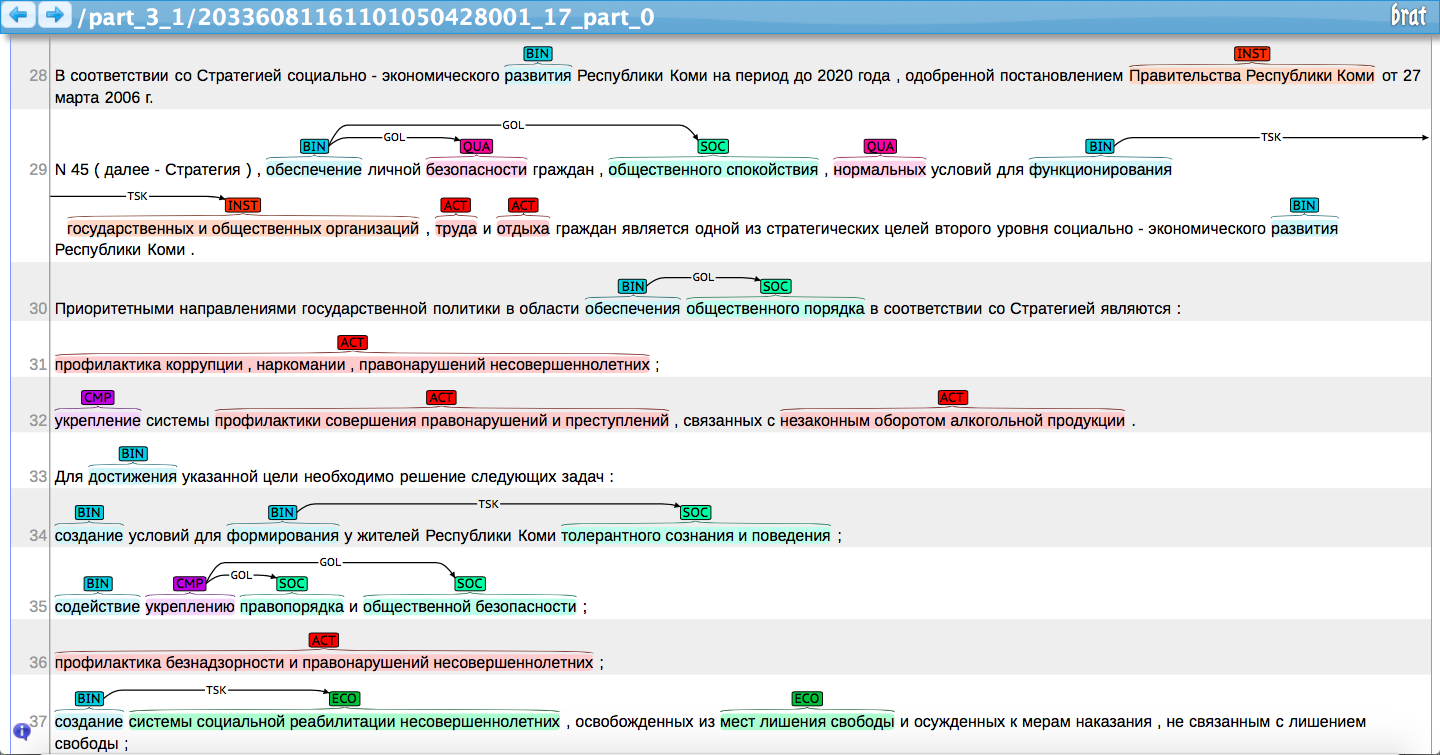}
    \caption{Annotation interface for assigning entities and relations.}
    \label{fig:brat}
\end{figure}

We develop guidelines for entity and relation identification in order to maintain uniformity of annotation in our corpus~\cite{rurebus}. Figure \ref{fig:brat} presents annotation interface for assigning entities and relations. 

We define eight types of entities: 
\begin{inparaenum}[1.]
    \item \texttt{MET} (metric)
    \item \texttt{ECO} (economics) 
    \item \texttt{BIN} (binary)
    \item \texttt{CMP} (compare)
    \item \texttt{QUA} (qualitative) 
    \item \texttt{ACT} (activity)
    \item \texttt{INST} (institutions) 
    \item \texttt{SOC} (social)
\end{inparaenum}

These entities associated with 11 semantic relations of 5 types:
\begin{inparaenum}[1.]
    \item Current situation: 
    \begin{inparaenum}
        \item Negative \texttt{NNG} (Now NeGative)
        \item Neutral \texttt{NNT} (Now NeutTral)
        \item Positive \texttt{NPS} (Now PoSitive)
    \end{inparaenum}
    \item Implemented changes/results (about the past):
    \begin{inparaenum}
        \item Negative \texttt{PNG} (Past Negative)
        \item Neutral \texttt{PNT} (Past NeutTral)
        \item Positive \texttt{PPS} (Past PoSitive)
    \end{inparaenum}
    \item Forecasts:
    \begin{inparaenum}
        \item Negative \texttt{FNG} (Future NeGative)
        \item Neutral \texttt{FNT} (Future NeutTral)
        \item Positive \texttt{FPS} (Future PoSitive)
    \end{inparaenum}
    \item \texttt{GOL} (abstract goals) 
    \item  \texttt{TSK} (specific tasks)
\end{inparaenum}

All annotations were obtained using a Brat Rapid Annotation Tool (BRAT) \cite{stenetorp2012brat}. Each document in the corpus was annotated by two annotators independently, while disagreements were resolved by a moderator. All annotation instructions are available at the \href{https://github.com/dialogue-evaluation/RuREBus}{GitHub repository}\footnote{https://github.com/dialogue-evaluation/RuREBus}. Below we present description of entity types. 

\subsection{Entity Descriptions}

\begin{comment}
\begin{table}[]
\centering
\begin{tabular}{|l|l|l|}
\hline
entity & example (eng)                      & example (ru)                  \\ \hline
\texttt{BIN}    & creation                           & создание                      \\ \hline
\texttt{MET}    & potato crop yield                       & урожайность картофеля         \\ \hline
\texttt{QUA}    & in a state of disrepair            & в аварийном состоянии         \\ \hline
\texttt{CMP}    & increased                          & увеличился                    \\ \hline
\texttt{SOC}    & public security                    & общественная безопасность     \\ \hline
\texttt{INST}   & Government of Komi Republic        & Правительство Республики Коми \\ \hline
\texttt{ECO}    & private plots                      & личных подсобных хозяйств     \\ \hline
\texttt{ACT}    &  drug monitoring  & мониторинг нарко ситуации    \\ \hline
\end{tabular}
\bigskip
\caption{Examples of annotated entities}
\label{table:ne_exmpl}
\end{table}
\end{comment}

\texttt{BIN}
 is a one-time action or binary characteristic. These entities represents one-time events such as construction, development, stimulation, formation, implementation, acquisition, involvement, absence, diversification, modernization, etc.

\texttt{MET}  entity is a numerical indicator or object on which a comparison operation is defined. In particular, these entities often describe labor productivity, planned and actual values of indicators, seismicity of a territory, the probability of a violation, economic growth, the degree of deterioration of a building, etc.

\texttt{QUA} represents a quality characteristic. Annotators were asked to identify spans of texts such as \textit{high}, \textit{ineffective}, \textit{limited}, \textit{big}, \textit{weak}, \textit{safe} as \texttt{QUA} entities.

\texttt{CMP} represents a comparative characteristic. Annotators were asked to identify spans of texts as \texttt{CMP} entities, associated with increasing, saturation, increasing decrease, activation, exceeding indicators, positive dynamics, improvement, expansion.

\texttt{SOC} is an entity related to social rights or social amenities. \texttt{SOC} entities other describe country population, housing quality, social protection, leisure activities, historical heritage, folk art, terms related to social rights or social amenities, etc.

\texttt{INST}
 entities represent various institutions, structures and organizations. In particular, annotators were asked to mark cultural and leisure facilities, family and child support organizations, cultural center as \texttt{INST}.

\texttt{ECO} is defined as an economic entity or infrastructure object. Entities of this type are associated with biological resources, innovative potential, domestic market, regional economy, energy balance, budget financing, fishing fleet, roads, library and museum funds, etc.

\texttt{ACT} is an event or specific activity. These entities are often combined with \texttt{BIN}, e.g., \textit{launched an educational project}, where \textit{launched} is marked as \texttt{BIN} and an \textit{educational project} as \texttt{ACT}. Entities of this type are associated with events like drug addiction prevention, orphan prevention, educational projects, psychological assistance.

\subsection{Relations Descriptions}
\texttt{GOL} represents aims and goals of program. It is used to describe changes and objective that are expected to be achieved as the results of actions, proposed by the program.

\texttt{TSK} denotes concrete actions planned by the program. Main difference between \texttt{TSK} and \texttt{GOL} is that the later one describe ``what" the program aims to achieve and the first one state ``how" it will be done.

The other nine relations are designed to describe perceptions of the present, past and future state of affairs. Past relations (\texttt{PPS}, \texttt{PNG}, \texttt{PNT}) describe the previous situation. Respectively present relations (\texttt{NPS}, \texttt{NNG}, \texttt{NNT}) present current situation. Last triplet (\texttt{FPS}, \texttt{FNG}, \texttt{FNT}) predicts trends, metrics or consequences of the program in a long-term perspective.          

Table \ref{table:ne_exmpl} presents examples of annotated relations.

\begin{table}[]
\centering
\begin{tabular}{|l|l|l|}
\hline
\textbf{Entity} & \textbf{Example (ENG)} &   \textbf{Example (RU)}         \\
\hline
\texttt{GOL} & \texttt{CMP} improving \texttt{SOC} public health & \texttt{CMP} укрепление  \texttt{SOC} здоровья населения      
\\ \hline
\texttt{TSK} & \texttt{BIN} halting \texttt{ECO} drug trafficking & \texttt{BIN} пресечение  \texttt{ECO} нарко трафика      
\\ \hline
\texttt{FPS} & \texttt{CMP} reduction of \texttt{MET} mortality rate & \texttt{CMP} снижение  \texttt{MET} уровня смертности        
\\ \hline
\end{tabular}
\bigskip
\caption{Examples of annotated relations.}
\label{table:ne_exmpl}
\end{table}

\subsection{Active learning}

We also employ active learning technique \cite{DBLP:journals/corr/ShenYLKA17}. Previously we obtained a subset of our corpus marked with this set of named entities and relations. Then we trained NER model and use it to markup unlabeled documents. Than documents were edited by annotators and verified by moderators. After that obtaining new part of final corpus model were retrained with this part added to training set.

In this work we employ NER model, namely char-CNN-BiLSTM-CRF (proposed by Lample et al \cite{lample-etal-2016} and further developed by Ma and Hovy \cite{ma2016endtoend}). This architecture is widely used as a robust baseline in sequence tagging tasks. We use FastText \cite{bojanowski2017enriching} embeddings trained by RusVectores \cite{KutuzovKuzmenko2017}. For relation extraction we also employ morphological, syntactical and semantical features, obtained from Compreno \cite{ZuyevK2013StatiSticalMT,anisimovich2012} and some hand-made features, such as capitalization templates and dependency tree distance between relation members.

\section{How to utilize named entities} 

In this and the following sections we provide an in-depth analysis of the annotated corpus and showcase  applications of textual analysis, based on the proposed annotation schema. We start with the description of annotated entities and provide insights into strategic planning based on entity-level analysis. Than we take the analysis to the next level and explore relations between entities and the way the relations help to structure information from strategic documents. 

\subsection{Basic statistics}

\begin{table}[ht]
% \arrayrulecolor[HTML]{DB5800}
% \centering
% \begin{table}[]
\begin{center}
\setlength{\tabcolsep}{4pt}
\begin{tabular}{|c|c|c|}
\hline
     & \textbf{Total} &\textbf{Mean len (std)} \\ \hline
\texttt{BIN}  & 14236 & 1.05 \space (0.28) \\ \hline
\texttt{MET}  & 6377  & 4.23 \space (3.50) \\ \hline
\texttt{QUA}  & 3611  & 1.14 \space (0.52) \\ \hline
\texttt{CMP}  & 4149  & 1.16 \space (0.78) \\ \hline
\texttt{SOC}  & 5037  & 2.77 \space (2.31) \\ \hline
\texttt{INST} & 3756  & 3.69 \space (2.81) \\ \hline
\texttt{ECO}  & 11422 & 2.78 \space (2.19) \\ \hline
\texttt{ACT}  & 5800  & 4.74 \space (4.57) \\ \hline
\end{tabular}
\end{center}
\caption{Statistics of annotated entities.}

% \end{table}
% \caption{NE counts}
\label{table:ta}
\end{table}

% \begin{table}
% \centering
% \begin{tabular}{|l|l|l|l|l|l|}

% \hline
% label & top 3 NE \\
% \hline
% \texttt{BIN} & реализации (1285), развитие (546), обеспечение (444),  \\ \hline
% \texttt{MET} & эффективности (129), энергетической эффективности (64), уровнем эффективности (60),  \\ \hline
% \texttt{QUA} & эффективного (69), эффективной (45), эффективное (43),  \\ \hline
% \texttt{CMP} & повышение (455), повышения (227), увеличение (196),  \\ \hline
% \texttt{SOC} & населения (50), граждан (49), образование (47),  \\ \hline
% \texttt{INST} & органов местного самоуправления (140), администрации (79), муниципального образования (48),  \\ \hline
% \texttt{ECO} & экономики (74), энергосбережения (73), бюджета (72),  \\ \hline
% \texttt{ACT} & мероприятий (103), территориального планирования (43), муниципальной программы (37),  \\ \hline

% \end{tabular}
% \bigskip
% \caption{Top 3 named entity}
% \label{tab:ne_top3}
% \end{table}

In this section we provide basic statistics based on annotated entities. There are 188 annotated documents in the training set, average number of named entities in document is 289, mean document length is 1787 tokens. All token-based statistics were obtained using 
\href{https://github.com/natasha/razdel}{razdel tokenizer.}\footnote{https://github.com/natasha/razdel}
\bigskip

Named entity types are highly imbalanced, which may lead to significant problems when training classifier. However, we believe that proposed corpus design represents real-life situation well. These difficulties should inspire researches to invent more sophisticated solutions, rather than prevent them from approaching the task.

\subsection{Named entity clustering} 
The main part of RuREBus are annotations of named entities. The types of the entities (such as `activity' or `institution') are quite broad to perform strategic analysis and planning. Clustering of entities into fine-grained subsets of entities that represent some concept, could be more useful in specific practical applications. Therefore, in this section we show how to use a simple technique  to investigate find semantically related subgroups of named entities. 
For instance, a cluster of entities may represent a specific social-oriented measures (such as `prevention of drug usage in youth'). We demonstrate how to use modern natural language processing methods to find semantic clusters of annotations.

The clustering procedure consists of the following steps. First, we preprocess texts: all textual representations of entities are lowercased and duplicates are removed. Then, we represent each entity with a vector, or embedding.  Finally, we applied $k$-means algorithm to find clusters.

The steps above a applied to each entity type separately. In the second step we represent each named entity with a vector. The vector (embedding) for each named entity has 1024 numbers. The vector representations for named entities were calculated using a combination of the FastText model pre-trained on Russian Wikipedia, and a bidirectional LSTM model (both models are implemented in the Flair library~\cite{akbik2018coling}). 

In the last step we use embeddings to find clusters. This step is performed by the $k$-means clustering algorithm. Number of clusters is a very important parameter; it varies dramatically from one type of entity to another. For example, entities of the \texttt{QUA} and \texttt{CMP} types have fewer subgroups variants. Hence, a reasonable number of clusters ($N$) for these types is smaller ($N=10$) than for \texttt{ECO} or \texttt{MET} ($N=50$). 
To select number of clusters we use classical silhouette analysis~\cite{rousseeuw1987silhouettes}; $k$-means clustering was performed by open-source library (scikit-learn).

\begin{figure}
    \centering
    \includegraphics[width=120mm]{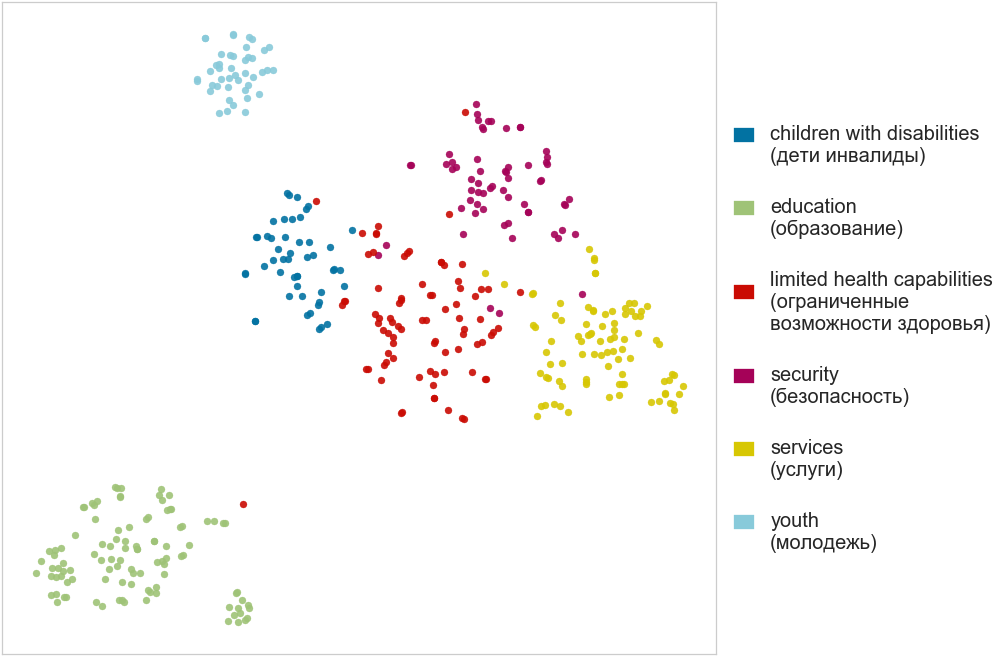}
    \caption{Projection of 6 clusters obtained for the \texttt{SOC} type (each color represents a cluster of named entities).}
    \label{Fig:1_images}
\end{figure}

In the rest of the section we briefly discuss the results of clustering. To represent results, we make use of a visualisation of a small fraction of the clusters (only 6 clusters are show in Figure \ref{Fig:1_images}). One can see that entities have been clustered in different groups.

We analyzed in details clusters derived within the \texttt{SOC} type. These clusters have complex structures, e.g. they contain closely related entities (`youth', `health of youth', `patriotic education of youth'), hierarchies of entities (`family', `youth family', `single parent youth family') as well as subgroups of opposite entities (`unity of the Russian nation' and `separatism'). This gives an interesting insight on how regions view the concept of youth and related entities in their strategic programs. Moreover, one can use the clusters to separate different subtypes of relations, e.g. goals (\texttt{GOL}) related to the `Family` cluster in a specific region of Russia. The following list represents an example of top entities in a cluster related to `Youth':

\begin{itemize}
\item \it{``patriotic education of children and youth'' (патриотического воспитания детей и молодежи);
\item ``self-realization of children and youth'' (самореализации детей и молодежи); 
\item ``education of the younger generation and youth'' (воспитания подрастающего поколения и молодежи); 
\item ``self-realization of youth'' (самореализации молодежи}).
\end{itemize}

Similar structures can be found in other typical clusters that make the whole corpus a very interesting resource for social, economic and geographic studies.

Finally, we list types of named entities along with clusters, which were found. The clusters clearly correspond to the intended meaning of the entity types (Table \ref{table:ne_exmpl}) and enable a detailed analysis of the annotated documents. 

\begin{table}[!htb]
\begin{center}
\begin{tabular}{|c|l|}
\hline
\textbf{Entity} & \textbf{Cluster names} \\ 
\hline
\texttt{BIN} & Done, Impossible, Negotiation, Creation, Improvement, Change, Necessity, ...   \\ \hline
\texttt{ACT} &  Programs and Events, Management, Organization, Support, Repair, ...  \\ \hline
\texttt{MET} & KPIs, Quantity, Effectiveness, Extent, Level, ... \\ \hline
\texttt{QUA} & Positive, Negative, Insufficient, Significant, Redundant, ...  \\ \hline
\texttt{CMP} & Increasing, Decreasing, More than / Less then, Negative dynamics, ... \\ \hline
\texttt{SOC} & Demographic trends, Education, Science, Culture, Sport, Family, ... \\ \hline
\texttt{INST} & Enterprises, Departments, Regions, Executive authorities, ...  \\ \hline
\texttt{ECO} & Industries, Innovations, Budgets, Taxes, Infrastructure, Energy, ...  \\ 
\hline
\end{tabular}
\end{center}
\caption{Typical clusters for entity types.}
\label{table:clusters}
\end{table}

\subsection{What \texttt{actions} are being taken?}
\texttt{ACT} is an entity that describes what actions should be taken in order to complete the tasks and to reach the goals. We can think of two scenarios for action-based analysis. First, we presume that all actions are planned more or less in the same fashion by different regions, as there are a lot of common goals. However, there might be some unique actions, such as ``creation of Cossack youth centers'' (\textit{создание казачьих молодежных центров}), which either reveal some specific needs of the region or unreasonable expenses. 
Second, we can estimate the cost of actions, based on data of previous years. This will enable on the fly evaluation of the strategic program budget.

\section{How to utilize relation}

\subsection{Basic statistics} 

In Table \ref{tab:rel_basics}, one can observe a number of relation occurrences based on the training set. Average amount of relations in document is 67. We also calculated a mean number of tokens between named entities spans participated in relation and the result is 0 for almost all relation types.  

\begin{table}[h]
\centering
\setlength{\tabcolsep}{4pt}
\begin{tabular}{|c|c|c|c|c|c|}
\hline
\textbf{RE}     & \textbf{Total} & \textbf{RE}     & \textbf{Total} & \textbf{RE}     & \textbf{Total} \\  \hline 
\texttt{GOL} & 3563 & \texttt{TSK} & 4613 &   & \\ \hline
\texttt{NPS} & 755  & \texttt{NNG} & 844 & \texttt{NNT} & 534  \\ \hline
\texttt{PPS} & 528  & \texttt{PNG} & 84  & \texttt{PNT} & 190  \\ \hline
\texttt{FPS} & 1167 & \texttt{FNG} & 229 & \texttt{FNT} & 141  \\ \hline
\end{tabular}
\bigskip
\caption{Relation statistics.}
\label{tab:rel_basics}
\end{table}

\subsection{Is change always good?} 

Some types of relations allow us to evaluate ongoing changes.
Positive assessments of changes are expressed by \texttt{NPS} and \texttt{PPS} relations. For example, \texttt{NPS(CMP, MET)}: ``decrease'' ({\it снижение}) --  ``gas prices'' ({\it цен на бензин}) or \texttt{PPS(CMP, MET)}: ``increased'' ({\it увеличивались}) --  ``cash income of the population'' ({\it денежные доходы населения}).

Negative assessments are expressed by \texttt{NNG} and \texttt{PNG} relations. For example, \texttt{NNG(MET, QUA)}: ``the housing cost'' ({\it стоимость жилья}) --  ``high'' ({\it высокая}) or \texttt{PNG(MET, CMP)}: ``the population size'' ({\it численность населения}) --  ``decreased'' ({\it сократилась}).

Neutral assessments of changes are expressed by \texttt{NNT} and \texttt{PNT} relations. For example, \texttt{NNT(SOC, QUA)}: ``quality of life'' ({\it качество жизни}) --  ``fair'' ({\it удовле-творительное}) or \texttt{PNT(BIN, ECO)}: ``the investment project'' ({\it инвестиционный проект}) --  ``developed'' ({\it разработан}).

The texts of the collection contain assessments of qualitative changes (i.e. a situation is compared between the past and the present) or assessments of the current state of affairs without comparison with the past. Therefore, entities involved in these relations are usually of \texttt{CMP} or \texttt{QUA} type.

An analysis of such relations could be useful for strategic planning of social and economic development of the country's regions. Such relations make it possible to judge how the implemented changes actually affect the life of society. However, it should be noted that assessments are subjective, and do not always coincide with the conventional wisdom.

\subsection{Do the tasks meet the goals?} 

Goals and tasks necessary to achieve the goals are expressed as relations between entities. We consider binary relations only, which allow to relate two entities. For example, a \texttt{goal} can be expressed as a relation between a \texttt{CMP} entity and a \texttt{MET} entity: ``improvement'' ({\it повышение}) --  ``accessibility of transport'' ({\it доступность транспорта}). A \texttt{task} can be expressed as a relation between a \texttt{BIN} entity and an \texttt{ECO} entity: ``commissioning'' ({\it ввод}) -- ``new metro lines'' ({\it новые линии метро}).

The presence of goals and tasks, expressed as fragments of text, allows us to measure the similarity between them. Different similarity types can be considered:
\begin{enumerate}
    \item co-occurrence frequency: if a goal and a task are frequently used in the same documents, there is a strong association between them. 
    \item semantic similarity: if a goal and a task consists of words, that share similar meaning, such as ``transport'' ({\it  транспорт}) and ``metro lines'' ({\it линии метро}), there is a semantic association between them.
    \item topic similarity: if a goal and a task belong to the same topic, such as  the goal ``road development'' ({\it развитие дорожной сети}) and the task ``reduction in the number of road accidents'' ({\it снижение числа дорожно-транспортных происшествий}) belong to the same topic, related to ``transport''.
\end{enumerate}
 
Measuring similarity helps to reveal whether the goal was split into tasks reasonably. If a there no tasks, similar to the stated goal, the achievement of this goal in practice becomes unlikely. The opposite might be the case, too: the absence of similar tasks reflects unrealistic goals. 

At the same time, being able to extract all tasks, may help to group them according to similarity measures, to find similar or even overlapping tasks and than to order them according to their complexity or urgency. 

We can employ the similarity measures mentioned above to align goals and tasks declared by federal and municipal subjects. The goal declared by a federal subject may be supported by smaller goal declared by its subdivisions, namely, municipal subjects. Although it is not necessary for all municipal subjects to share goals, the absence of common goals can reveal potential managerial and administrative weaknesses. 

At the same time goals declared by municipal subjects should follow the main development direction. If the average similarity between the goals declared on different levels is low, it means that the region lacks coherent coordination of planning authorities. 

To conclude with, the analysis of \texttt{goal} and \texttt{task} relations can be used in several ways. It can be applied both to a single strategic document and to multiple strategic documents, prepared in a region. In the first case, the relation analysis can help to structure goal setting along with goal decomposition and task prioritization. In the second case the relation analysis allows to discover coherence problems between different levels of subdivisions. 

\subsection{Temporal analysis of \texttt{past} and \texttt{present} relations}
The time component of extracted relations (current state / implemented changes / forecasts) allows us to measure the proportion of the work done to the planned work, in other words, understand whether a document contains a report on the work done rather than a plan for future work. Such a simple metric can monitor whether there is any success in reaching goals within a region over years, or documents contain only plans.

\begin{table}[t]
\begin{center}
 \begin{tabular}{|c|c|} 
 \hline
 \textbf{Republic, Region} &\textbf{ Done work to plans ratio} \\ [0.5ex] 
 \hline
 %\hline
 Komi Republic, Pechora Municipal District  & 0.596  \\ 
 \hline
 Ryazan Region, Mikhailovsky Municipal District & 0.523 \\
 \hline
 ... & ... \\
 \hline
 Voronezh Region, Khlebenskoe & 0.06 \\
 \hline
 Moscow Region, Dmitrovsky & 0.02 \\
 \hline
\end{tabular}
\end{center}
\caption{Analysis of done work to planned ratio analysis.}
\end{table}

Furthermore, we can match forecasts from previous years' plans with the descriptions of implemented changes of current year to check whether the plans were implemented or not.

Altogether, we can automatically calculate the percentage of goals stated in previous years and achieved this year. In addition, we could observe the tendencies in some key metrics over time and regions from current state relations, for example, how the youth crime level has changed in some region during the last 5 years.

\section{Conclusion}
The exponential growth in the volume of information overwhelms many domains of human activity, including state regulation and planning. As government documents exist in the form of written text, the role of language technology is increasingly important. In this paper we showcase two well-known tasks, named entity recognition (NER) and relation extraction (RE), formulated for the strategic planning domain. 

In this on-going project we intend to carry out the full cycle of language technology development: we start from raw texts, elaborate an annotation schema, annotate hundreds of documents with the help of human-in-the-loop approach, train domain-specific models for the tasks under consideration. Finally, we design a few analytical applications, which demonstrate the relevance and validity of the designed language resource.  Not only we managed to implement the whole NLP pipeline for a novel application, but also we have shown that governmental documents can be subjected to computational analysis. Future research of strategic planning and e-Goverment would be able to benefit from the developed methods and tools as we release all results and code in open access. These cab be used to extract knowledge and gain insights from strategic planning documents or can be applied to other domains.

The future work directions reflect the rapid development of language technology. A large language model may be trained on the strategic documents and enhance the quality of downstream tasks, NER and RE. The project may benefit from recent cross-lingual methods as this would allow to conduct comparison between strategic planning in different countries. 

%\paragraph{Acknowledgements}
\textbf{Acknowledgements}
Work on corpus annotation and manuscript was carried out by Ekaterina Artemova, Elena Tutubalina, and Veronika Sarkisyan and was funded by the framework of the HSE University Basic Research Program and Russian Academic Excellence Project ``5-100''. Work on annotation of the part of the corpus was carried out by Tatiana Batura and was funded by RFBR according to the research project N 19-07-01134.

\bibliography{references} 
\bibliographystyle{splncs04}

\end{document}